\documentclass[10pt,conference]{IEEEtran}
\IEEEoverridecommandlockouts

\usepackage{cite}
\usepackage{amsmath,amssymb,amsfonts,amsthm}
\usepackage[linesnumbered,ruled]{algorithm2e}
\usepackage{enumitem}
\usepackage{graphicx}
\usepackage{textcomp}
\usepackage[dvipsnames,table]{xcolor}
\usepackage[colorinlistoftodos,prependcaption,obeyFinal]{todonotes}
\usepackage{colortbl}
\usepackage{verbatim}

\usepackage{kotex}
\usepackage{algpseudocode}
\usepackage{booktabs}
\usepackage{tabularx}
\usepackage{multirow}
\usepackage{comment}
\usepackage{pgfplots}
\pgfplotsset{compat=1.10}
\usepackage{subfigure}
\usepackage[most]{tcolorbox}
\usepackage{arydshln}
\usepackage{colortbl}
\usepackage{pifont}
\usepackage{url}

\usepackage{booktabs}
\usepackage{pifont}

\def\BibTeX{{\rm B\kern-.05em{\sc i\kern-.025em b}\kern-.08em
    T\kern-.1667em\lower.7ex\hbox{E}\kern-.125emX}}

\definecolor{ruby}{RGB}{242,0,20}
\definecolor{emerald}{RGB}{26,121,42}
\definecolor{topaz}{RGB}{236,185,57}
\definecolor{sapphire}{RGB}{14,26,164}
\definecolor{flax}{rgb}{0.93, 0.86, 0.51}

\newtheorem{example}{Example}

\newcommand{\tuple}[1]{{\langle{#1}\rangle}}
\newcommand{\cmark}{\ding{51}}
\newcommand{\greencmark}{\textcolor{ForestGreen}{\cmark}}
\newcommand{\xmark}{\ding{55}}
\newcommand{\redxmark}{\textcolor{BrickRed}{\xmark}}

\newcolumntype{Y}{>{\centering\arraybackslash}X}

\presetkeys{todonotes}{inline, color=topaz!40, bordercolor=topaz}{}

\newcommand{\sysname}{\textsc{Sage}\xspace}

\newcommand{\LLM}{\textnormal{\textsf{LLM}}}
\newcommand{\GT}{\textnormal{\textsf{GT}}}
\newcommand{\textsymb}[1]{\textnormal{\texttt{#1}}}

\colorlet{promptbg}{gray!80}
\newtcolorbox{promptbox}[1][]{
  colback=promptbg!10,
  colframe=promptbg,
  coltitle=black,
  colbacktitle=promptbg!40,
  boxrule=0.0pt,
  arc=2pt,
  left=2pt,
  right=2pt,
  top=2pt,
  bottom=2pt,
  fonttitle=\bfseries,
  breakable,
  title=Prompt,
  #1
}

\SetKwInput{KwInput}{Input}
\SetKwInput{KwOutput}{Output}

\usetikzlibrary{calc}

\begin{document}

\title{\sysname: Specification-Aware Grammar Extraction for Automated Test Case Generation with LLMs}

\author{
\IEEEauthorblockN{
    Aditi\IEEEauthorrefmark{1},
    Hyunwoo Park\IEEEauthorrefmark{1},
    Sicheol Sung\IEEEauthorrefmark{2},
    Yo-Sub Han\IEEEauthorrefmark{2},
    Sang-Ki Ko\IEEEauthorrefmark{1}\thanks{Corresponding author}
}
\IEEEauthorblockA{\IEEEauthorrefmark{1}University of Seoul, South Korea}
\IEEEauthorblockA{\IEEEauthorrefmark{2}Yonsei University, South Korea}
\IEEEauthorblockA{
    Email: \{aditimzu16, sangkiko\}@uos.ac.kr,
    illuminare0@gmail.com, \{sicheol.sung, emmous\}@yonsei.ac.kr
}
}

\maketitle

\begin{abstract}


Grammar-based test case generation has proven effective for competitive programming problems, but generating valid and general grammars from natural language specifications remains a key challenge, especially under limited supervision. Context-Free Grammars with Counters (CCFGs) have recently been introduced as a formalism to represent such specifications with logical constraints by storing and reusing counter values during derivation. In this work, we explore the use of open-source large language models (LLMs) to induce CCFGs from specifications using a small number of labeled examples and verifiable reward-guided reinforcement learning. Our approach first fine-tunes an open-source LLM to perform specification-to-grammar translation, and further applies Group Relative Policy Optimization (GRPO) to enhance grammar validity and generality. We also examine the effectiveness of iterative feedback for open and closed-source LLMs in correcting syntactic and semantic errors in generated grammars. 

Experimental results show that our approach (\sysname) achieves stronger generalization and outperforms 17 open and closed-source LLMs in both grammar quality and test effectiveness, improving over the state-of-the-art by 15.92\%p in grammar validity and 12.34\%p in test effectiveness. 
We provide our implementation and dataset at the following anonymous repository:
\url{https://anonymous.4open.science/r/SAGE-5714}.
\end{abstract}



\section{Introduction}


Automated Test Case Generation (ATCG)\cite{AnandBCCCGHHMOE13} plays a pivotal role in software engineering, particularly within competitive programming, where the performance and correctness of an algorithm are rigorously tested against a wide range of complex inputs. As software systems and algorithmic challenges grow increasingly sophisticated, it becomes impractical to craft test cases manually. This leads people to adopt automated approaches. The development of large language models~(LLMs) has significantly advanced the efficient and scalable generation of test cases. However, these models lack interpretation of complex input specifications and the reliable identification of edge cases, often requiring supplementary manual effort to ensure comprehensive coverage. 

The inadequacy of test suites has been a persistent issue in software testing.
For example, the Defects4J dataset, as highlighted
by~\cite{FraserA11}, has shown that incomplete test cases lead
to insufficient program analysis. Similarly, the CodeNet dataset, a benchmark
for competitive programming~\cite{Puri0JZDZD0CDTB21}, suffers
from a lack of high-quality test cases, as noted
by~\cite{ZhaoHMLZJLZ024}. These limitations not only hinder robust
algorithm validation but also exacerbate challenges in program
repair~\cite{TianLPKLHKB22}. Even recent prompt-based
methods leveraging LLMs, such as ChatGPT, and
mutation-based frameworks like MuTAP~\cite{DakhelNMKD24}, often produce invalid
or incomplete test cases for competitive programming due to the complexity of
test case specifications.  

These shortcomings point to a crucial weakness in automated testing: existing approaches struggle to provide test cases that rigorously follow intricate problem constraints while capturing edge situations that are necessary for algorithm validation. For example, errors in certain program synthesis benchmarks, as disclosed by~\cite{LiuXW023}, highlight the need for strong testing frameworks that can fill up these gaps. 

To tackle these challenges, we utilized Context-Free
Grammars with Counters~(CCFGs), a novel formalism
that integrates both syntactic and semantic elements of
problem input specifications~\cite{sung2025logicase} to generate test cases using LLMs. 
One of the central challenges in applying LLMs to ATCG for complex competitive programming lies in their limited
ability to accurately capture and utilize the syntactic structure of CCFGs.
These grammars are crucial for generating valid and semantically meaningful test
cases, yet most LLMs struggle to model their hierarchical and recursive nature,
often producing syntactically incorrect or structurally incoherent outputs. This
syntactic fragility severely limits the reliability and effectiveness of
LLM-based test case generation.

Compounding this issue is the high cost---both computational and financial---of leveraging commercial LLM APIs. These models, while powerful, often require expensive access fees and substantial computational resources, making them impractical for many academic institutions, independent researchers, or organizations with limited budgets. Moreover, most commercial models are closed-source, restricting transparency, customization, and reproducibility.

Another challenge is the one-shot or few-shot limitation of many LLMs in grammar generation. Even when guided by instructions or examples, these models frequently fail to produce valid grammars without external feedback or correction. 
These limitations underscore the need for approaches that are not only
syntactically-aware and cost-effective, but also capable of iterative refinement and learning from weak supervision. Recent studies have shown that
self-repair strategies lead to greater performance gains compared to single-turn
sampling~\cite{instruction_finetuning,DBLP:conf/iclr/NijkampPHTWZSX23,DBLP:conf/iclr/ZhengDGCNS25}.
\begin{figure*}
    \centering
    \includegraphics[width=1.\linewidth,clip,trim=0 0.3cm 0 0]{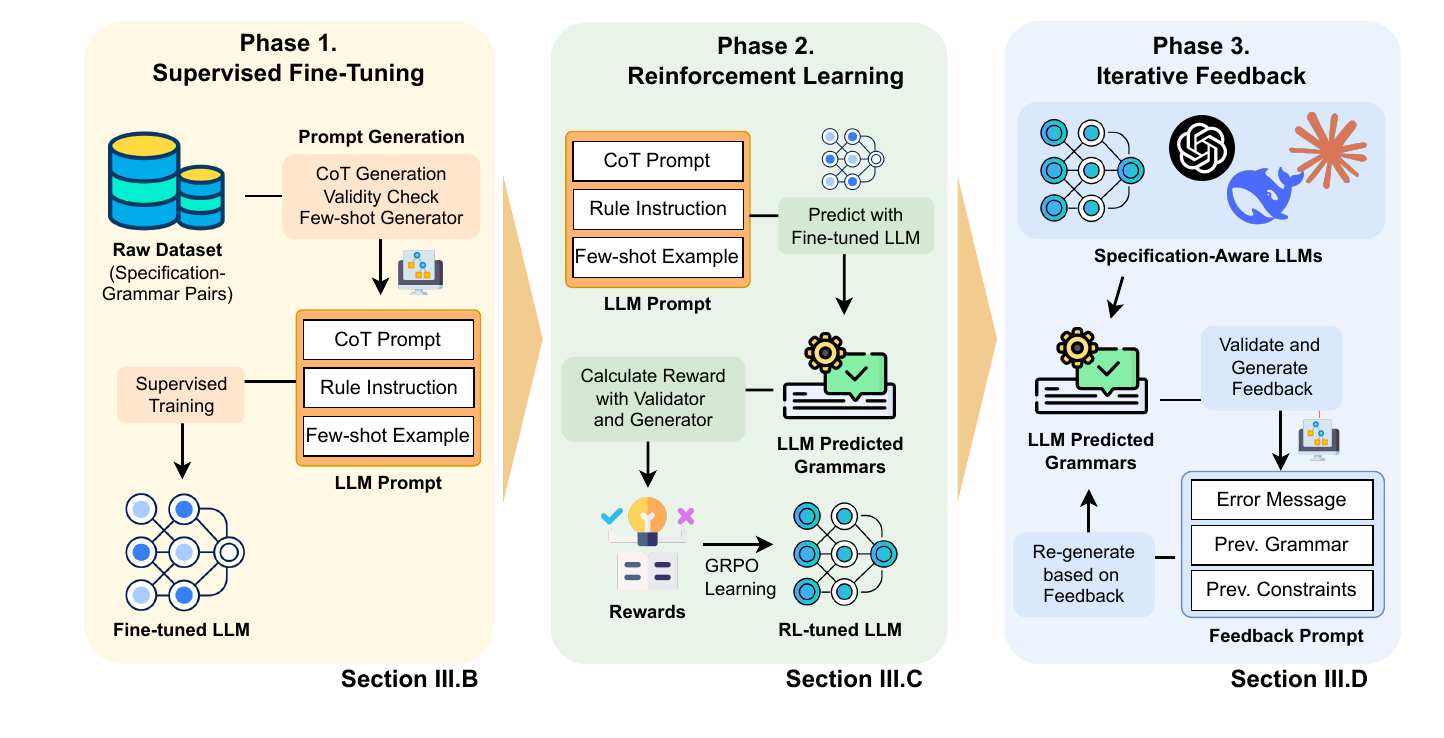}
    \caption{Overview of the proposed framework}
    \label{fig:overview}
\end{figure*}

We propose the \sysname, a \underline{S}pecification-\underline{A}ware
\underline{G}rammar \underline{E}xtraction to address these limitations.
\sysname is a framework based on supervised fine-tuning and reinforcement
learning~(RL) with grammar-based verifiable rewards on an open-source LLM. To
prove the effectiveness of the \sysname~ framework, we make the
following contributions:

\begin{enumerate}
\item \textbf{Supervised Fine-Tuning for Specification-to-Grammar Translation}:
We fine-tune the DeepSeek-R1-Distill-Qwen-14B model using supervised learning
to enhance the validity and generality of CCFGs. We achieve up to 78\%
effectiveness in test case generation, significantly surpassing baselines including closed-source LLMs.

\item \textbf{RL for Specification-Aware Grammar Generation}:
We further apply RL to the fine-tuned base model by
explicitly rewarding grammars that generate valid and general test cases. 

\item \textbf{Multi-Turn Feedback Iteration for Closed LLMs}:
We introduce a multi-turn feedback framework that allows the model to refine grammar outputs across successive prompts iteratively. This approach significantly improves syntactic validity and robustness for closed-source LLMs due to their reasoning capability.

\item \textbf{Benchmarking Various LLMs}:
We demonstrate that our approach with an open-source LLM outperforms 17~LLMs spanning several orders of magnitude in size. 
\end{enumerate}


We address the core challenges in grammar-driven ATCG and close the performance
gap between open and closed models by formulating the following research
questions:


\begin{tcolorbox}[boxsep=1pt,left=4pt,right=4pt,top=4pt,bottom=4pt,boxrule=0pt]
\textbf{RQ1:} Can we induce high-quality input grammars from limited specification-to-grammar supervision?
\end{tcolorbox}

\begin{tcolorbox}[boxsep=1pt,left=4pt,right=4pt,top=4pt,bottom=4pt,boxrule=0pt]
\textbf{RQ2:} Does RL with verifiable rewards improve the validity and generality of induced grammars?
\end{tcolorbox}

\begin{tcolorbox}[boxsep=1pt,left=4pt,right=4pt,top=4pt,bottom=4pt,boxrule=0pt]
\textbf{RQ3:} Does iterative feedback improve grammar quality over single-turn generation for closed-source models?
\end{tcolorbox}


\section{Background}

\subsection{Context-Free Grammars with Counters}



A key difficulty of ATCG lies in producing
inputs that not only adhere to syntactic rules but also satisfy structural and
numerical constraints implicit in input specifications. Classical Context-Free
Grammars~(CFGs) describe input formats through non-terminals and production
rules. CFGs have been widely used to enable effective fuzzy testing by ensuring
syntactic validity of generated test cases~\cite{SrivastavaP21}. However, CFGs
often fall short when the specification includes interdependent
constraints---such as length counts or value consistency---since the derivations
of non-terminals do not influence one another.


This limitation has motivated the development of Context-Free Grammars with
Counters~(CCFGs), which extend CFGs by incorporating logical and numerical
constraints into the derivation process~\cite{sung2025logicase}. The use of
counters within production rules enables grammars to access intermediate values
generated during derivation, allowing subsequent derivations to reflect
constraints imposed by earlier values.

Example~\ref{example:spec-ccfg} presents an input specification for the
problem~``1369\_C. RationalLee'' from Codeforces, along with the corresponding
CCFG production rules derived from the specification, as shown in
\cite{sung2025logicase}.

\begin{example}[1369\_C. RationalLee from Codeforces]\hfill
\label{example:spec-ccfg}
\begin{itemize}
\item
The first line contains one integer~$t$. \\
$\blacktriangleright S \to [t]\ \textsymb{<n>}\ T_t$

\item
Next $3t$ lines contain test cases---one per three lines. \\
The first line contains two integers~$n$ and~$k$. \\
$\blacktriangleright T_i \to T_{i-1}\ \textsymb{<n>}\ [n]\ \textsymb{<s>}\ [k]\ \textsymb{<n>}\ L_n\ \textsymb{<n>}\ Z_k$ \\
$\blacktriangleright T_1 \to [n]\ \textsymb{<s>}\ [k]\ \textsymb{<n>}\ L_n\ \textsymb{<n>}\ Z_k$

\item
The second line contains $n$~integers~$a_1, \ldots, a_n$. \\
$\blacktriangleright  L_i \to L_{i-1}\ \textsymb{<s>} a_i$ \quad
$\blacktriangleright L_1 \to a_1$

\item
The third line contains $k$~integers~$b_1, \ldots, b_k$. \\
$\blacktriangleright Z_i \to Z_{i-1}\ \textsymb{<s>}\ b_i$ \quad
$\blacktriangleright Z_1 \to b_1$
\end{itemize}
\end{example}

Each line of the input specification in Example~\ref{example:spec-ccfg}
corresponds directly to a nonterminal in the corresponding CCFG: the overall
structure is captured by~$S$, repeated test cases by~$T$, and the sequences of
integers by~$L$ and $Z$. Since each production rule is parameterized by a
counter~$i$, it derives the exact number of repetitions specified by~$t, n$
or~$k$.


\subsection{Validity and Generality in Grammar-Based Generation}
\label{subsection:validity-generality}



We say that a test case is valid if it conforms to the input specification. We
use two key properties to evaluate the quality of a grammar for a given input
specification: \emph{validity} and \emph{generality}. Validity refers to whether
the test cases generated by the grammar are valid, while generality measures
whether the grammar can generate all valid test cases defined by the
specification.

\begin{figure}[h]
\centering
\includegraphics[width=\linewidth]{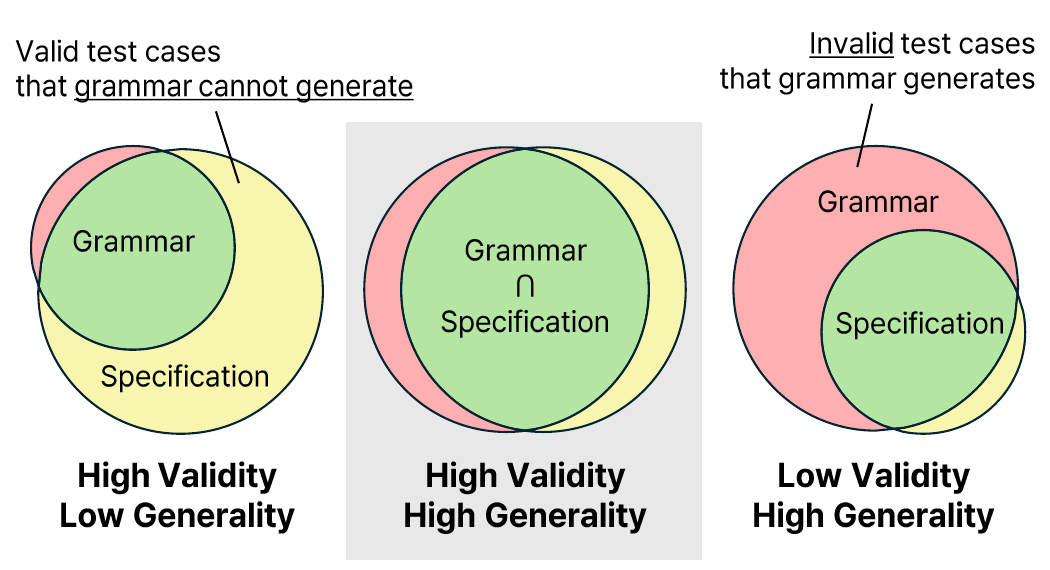}
\caption{Visualization of validity and generality scores. Set-based validity
is~$1$ if and only if the grammar does not generate any invalid test cases and
set-based generality is~$1$ if and only if the grammar can generate all valid
test cases.}
\label{fig:metric}
\end{figure}

\emph{Element-based} validity and generality evaluate an individual test case.
In contrast, \emph{set-based validity} evaluates whether all test cases
generated by the grammar is valid, and \emph{set-based generality} measures
whether the grammar can generate all known valid test cases. These metrics serve
a dual purpose: they guide the RL process as reward
functions and are also used to evaluate the quality of the grammars inferred by
models. Figure~\ref{fig:metric} illustrates the relationship between validity and
generality.

\subsection{Group Relative Policy Optimization (GRPO)}

To generate the valid and general grammars from natural language specifications, we adopt Group Relative Policy Optimization (GRPO)~\cite{grpo2024}, a reinforcement learning method adapted from Proximal Policy Optimization (PPO)~\cite{SchulmanWDRK17}. While PPO optimizes policy networks using clipped surrogate objectives to ensure stable updates, GRPO extends this idea to settings where optimization should account for group-level objectives---such as maximizing grammar validity and generality across a batch of problems or specification groups.

Formally, PPO updates the policy parameters~$\theta$ at each timestep~$t$ by
maximizing the following clipped objective:
\[
L^{\text{PPO}}(\theta) = \mathbb{E}_{t}\left[\min\left(r_t(\theta) \hat{A}_t, \ \text{clip}(r_t(\theta), 1 - \epsilon, 1 + \epsilon)\hat{A}_t\right) \right],
\]
where $
    r_t(\theta) = \frac{\pi_\theta(a_t|s_t)}{\pi_{\theta_{\text{old}}}(a_t|s_t)}$
is the probability ratio between the new policy~$\psi_\theta$ and the old
policy~$\psi_{\theta_\text{old}}$, and \(\hat{A}_t\) is the estimated advantage
at timestep~$t$.

GRPO generalizes this framework by aggregating advantage estimates across related task groups, enabling learning that is sensitive to structural similarity. Specifically, GRPO introduces a \textit{group-relative advantage}:
\[
\hat{A}^{(g)}_t = \hat{A}_t - \mathbb{E}_{t'\in g}[\hat{A}_{t'}],
\]
where $g$ denotes a group of related tasks or specifications, and the expectation is taken over the group. The GRPO objective then becomes:
\begin{flalign*}
& L^{\text{GRPO}}(\theta) & \\
& = \mathbb{E}_{g} \left[
    \mathbb{E}_{t \in g}\left[\min\left(r_t(\theta) \hat{A}^{(g)}_t, \ \text{clip}(r_t(\theta), 1{-}\epsilon, 1{+}\epsilon) \hat{A}^{(g)}_t\right)\right]
\right]. &
\end{flalign*}

In our setup, an open-source LLM fine-tuned on a small number of labeled (specification-to-CCFG) examples is further optimized using GRPO. Here, the policy corresponds to the model that generates grammar candidate where the reward function is computed based on the validity and generality of the generated grammar. 

GRPO encourages a form of structured exploration, where models are encouraged not only to produce valid grammars for individual instances, but also to generalize across structurally similar problem classes. This allows us to extend the benefits of grammar-based test case generation to settings where labeled grammars are scarce or unavailable.


\section{Methodology}

\subsection{Problem Setting and Objectives}


We focus on the task of ATCG for competitive programming problems, where input
formats are governed by complex syntactic and semantic constraints. The central
challenge lies in generating test inputs that are both syntactically valid and
semantically consistent with the problem specification. Our framework integrates
LLMs with a combination of instruction-based prompting, feedback-driven
refinement, and fine-tuning via supervised learning and reinforcement learning
methods. The objective is to induce high-quality grammars that generalize across
diverse problem specifications and support valid, diverse, and edge-covering
test case generation.

\subsection{Fine-Tuning LLM for Spec-to-Grammar Translation}

We generate test cases for a given input specification by translating the
specification into its corresponding CCFG. Commercial LLMs often struggle to generate complex CCFGs from specifications because they are not aligned for generating CCFGs. Therefore, we further train
the LLMs through Supervised Fine-Tuning~(SFT) to improve their performance.

To align LLMs with CCFG generation, we designed high-quality prompt-response pairs. The prompt consists of a role instruction, specification-CCFG pair examples, and an output instruction that dictates the format for the output. The response consists of CoT-CCFG pairs that match the needs of the output instruction. The LLMs construct the output sequences for the prompt and calculate the loss compared to the response using the Cross-entropy loss function. This way, the LLMs learn the appropriate knowledge to generate the appropriate CCFGs for the specification. 





\subsection{Validity and Generality-Guided RL}


We utilize the GRPO algorithm to improve the quality of LLM-generated CCFGs. We
define reward functions for the strategy based on element-based validity~($R_V$)
and element-based generality~($R_G$). Specifically, (1)~the element-based
validity of a grammar is defined as the ratio of valid test cases to the total
number of test cases generated by the grammar, and (2)~the element-based
generality is the ratio of test cases the grammar can parse to the number of
ground-truth test cases.

Algorithm~\ref{alg:reward_function} details how these scores are computed. The
validity and generality scores are estimated by sampling $k$~test cases from the
LLM-generated grammar~$G_\LLM$ and the ground-truth
grammar~$G_\GT$~(Lines~\ref{line:end_validity} and \ref{line:end_generality}),
respectively. The total score is then defined as a weighted sum of these
individual scores. These reward functions not only encourage the model to
generate CCFGs closely matching the ground truth,
but also emphasize its capacity
to self-evaluate.


\begin{algorithm}[htbp]
\caption{Calculating CCFG Reward for RL}
\label{alg:reward_function} 

\KwInput{Generated grammar~$G_\LLM$, ground truth~$G_\GT$}
\KwOutput{Reward score~$R$}

\lIf{$G_\LLM$ is not well-formed}{\Return $0$}

\label{line:begin_validity}
Sample a set~$T_\LLM$ of $k$~test cases from $G_\LLM$\;
$R_V \gets \#\{t \in T_\LLM : G_\GT \text{ can parse } t\} / k$\;
\Comment{Compute element-based validity~$R_V$}
\label{line:end_validity}

\label{line:begin_generality}
Sample a set~$T_\GT$ of $k$~test cases from $G_\GT$ \\
$R_G \gets \#\{t \in T_\GT : G_\LLM \text{ can parse } t\} / k$\;
\Comment{Compute element-based generality~$R_G$} \\
\label{line:end_generality}

\Return{$R_V \cdot R_G$}
\end{algorithm}

\subsection{Iterative Grammar Refinement}
The iterative grammar refinement enables us to progressively improve both the syntactic validity and generality of the generated grammars. By leveraging model feedback across multiple iterations---guided by test case generation using the CCFG generation parser---we are able to systematically correct initial errors, reinforce syntactically and semantically sound patterns, and align the model outputs more closely with the intricacies of complex specification requirements. This iterative process not only enhances the quality of the generated grammars but also increases their reliability for CCFG generation.
During this refinement, we observed that most errors fall into the following key categories:
\begin{itemize}
\item \textbf{Null Grammar Error}:
The model fails to generate productions or constraints, typically due to
extraction failure.

\item \textbf{Unbracketed Counter Variable Error}:
A counter variable is used in a constraint or production without proper
bracketing or definition, causing parsing failure.
    
\item \textbf{Missing Variable Reference Error}:
A variable is defined in constraints but not used in productions, or vice versa,
leading to inconsistency.
    
\item \textbf{Node Overflow Error}:
Raised when a production allows values far beyond the constraint bounds. For
example, a production uses `\texttt{[1-9][0-9][0-9]}' while the constraint is $1
\leq n \leq 10^9$.
    
\item \textbf{Invalid Non-terminal Error}:
Production rules contain invalid representations of non-terminals, such as
`\texttt{<A->>}' or `\texttt{<Test\_case>}'. Note that we use
underscore~`\texttt{\_}' to denote non-terminals with subscripts,
as in~`\texttt{<A\_i>}'.


\end{itemize}

\begin{figure}[htbp]
    \centering
    \includegraphics[width=1.0\linewidth,trim=0 1.5cm 0 0, clip]{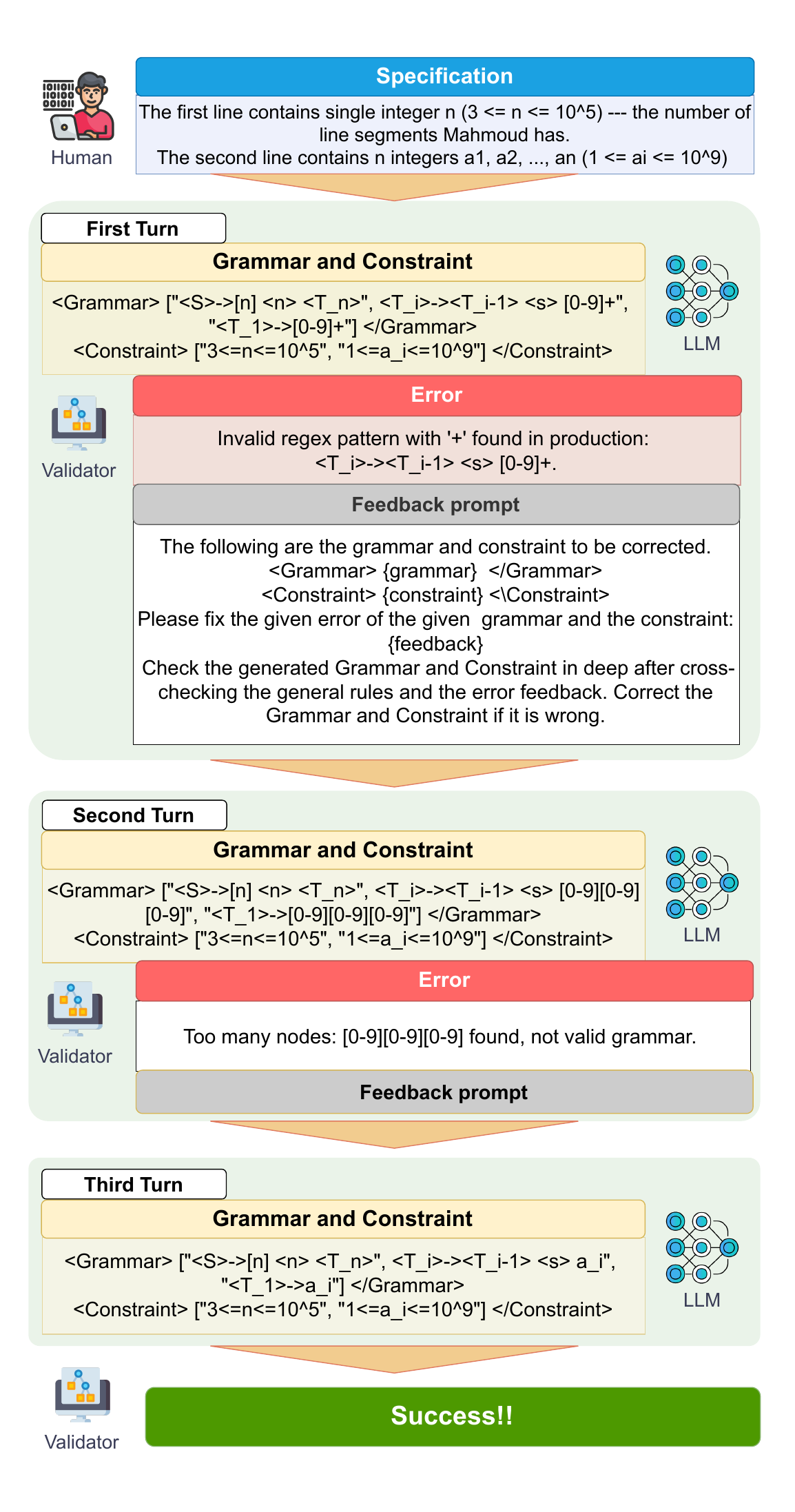}
    \caption{Overview of the iterative grammar refinement process.
    }
    \label{fig:iteration_overview}
\end{figure}

Figure~\ref{fig:iteration_overview} illustrates the overall process of iterative grammar refinement, which is bounded to a maximum of five iterations. This process is designed to incrementally enhance the quality of the generated grammars by systematically analyzing the types of errors present in each iteration. By identifying and categorizing these errors---such as incorrect production rules, constraint violations, or structural inconsistencies---we enable the application of targeted fixes tailored to specific failure modes. These corrective strategies are incorporated into each refinement cycle, allowing the model to learn from previous mistakes and produce increasingly robust and accurate CCFGs over time.

\subsection{CCFG-Based Test Case Generation}

A CCFG generates strings that satisfy given constraints by ensuring that
variable values sampled during derivation do not violate those constraints. In
the case of standard CFGs, one can sample a string uniformly at random in linear
time with respect to the output length by simulating the derivation process.
However, for CCFGs, no known polynomial-time algorithm exists for sampling
strings that satisfy arbitrary constraints, even in simple cases. This
complexity arises because checking whether a partial derivation can be completed
without violating future constraints may require backtracking or solving
constraint satisfaction problems, which is generally computationally hard.

Therefore, we adopt a Las Vegas approach to constrained sampling. During
derivation, we track the feasible value intervals of variables based on
accumulated constraints, and repeatedly sample values from these intervals until
all conditions are met. Although the algorithm may retry sampling multiple
times, it always returns a correct result when it terminates.

Algorithm~\ref{alg:generation} outlines the procedure. The algorithm extends the
string generation of CFGs by sampling and storing numeric values to satisfy
constraints~(Lines~\ref{line:begin_sample} and \ref{line:end_sample}) and
selecting production rule based on the variable
values~(Line~\ref{line:production}).

\begin{algorithm}[htbp]
\newcommand{\assign}{{\textsf{assign}}}
\caption{Sampling Test Cases from CCFG}
\label{alg:generation}
\KwInput{A CCFG~$G$}
\KwOutput{A test case~$x$}
$S_0 \gets \text{the initial nonterminal of the grammar~$G$}$\;
$x \gets \tuple{S_0}$; $A \gets \emptyset$\;
\While{there exists nonterminals or variables in $x$}{
    $t \gets \text{the leftmost nonterminal or variable of $x$}$\;
    \uIf{$t$ is a nonterminal~$X_i$}{
        \uIf{$i$ is a number~$k$}{
            Sample a production~$X_k \to \beta$ in $G$\; \label{line:production}
            Rewrite $t = X_k$ in $x$ with $\beta$\;
        }
        \ElseIf{$i$ is a variable~$v$}{
            Let $k$ be the value of $v$ in A\;
            Rewrite $u = X_v$ in $x$ with $X_k$\;
        }
    }
    \ElseIf{$t$ is a variable~$v$}{
        Sample the value~$k$ for $v$ satisfying constraints\;
        \label{line:begin_sample}
        $A \gets A \cup \{ v \mapsto k \}$\;
        \label{line:end_sample}
    }
}
\Return $x$
\end{algorithm}

\section{Experimental Setup}


We evaluate the practical usefulness of CCFGs by combining detailed prompt
designs with supervised fine-tuning and RL, validating their
effectiveness through experiments on diverse competitive programming tasks.

\subsection{Dataset}

We utilize the CodeContests dataset~\cite{LiAlphaCode22}, which includes a
variety of programming challenges from several competitive platforms. The
algorithms implemented in various programming languages are included in this
dataset, together with \emph{generated}, \emph{private}, and \emph{public} test
cases for most problems. Among the 1,200 problems with human-generated CCFGs
annotated by Sung et al.~\cite{sung2025logicase}, we use 960 for training and
240 for evaluation. To evaluate the effectiveness of our generated CCFGs against the test cases, we randomly sample 10 correct and 10 incorrect solutions for each specification.

\subsection{Base LLM}
  
We utilize the \textbf{DeepSeek-R1-Distill-Qwen-14B} model as our base LLM. To adapt the model for grammar-based test case generation, we apply SFT and RL with GRPO using structured
instruction-response pairs. Each prompt is carefully crafted with a combination
of \textit{CoT} reasoning and explicit \textit{instruction-based} guidance,
enabling the model to better capture the hierarchical structure of CCFGs.

\subsection{Hyperparameters}

For computing the validity and generality of grammars, we set the number of
sampled test cases in Algorithm~\ref{alg:reward_function} to $k = 5$. 
For reproducibility, all hyperparameters used in our paper
are listed in Table~\ref{tab:hyperparameter}.


\begin{table}[htbp]
\centering
\caption{Hyperparameters for training of large language models}
\label{tab:hyperparameter}
\subfigure[Supervised Fine-Tuning]{
\begin{tabular}{l@{\hspace{8pt}}l}
\toprule
\bf Hyperparameter & \bf Value \\
\midrule
Learning rate &
1e-5 \\ 
Learning rate scheduler & Cosine \\ 
Optimizer & AdamW \\ 
\bottomrule
\end{tabular}
}
\hfill
\subfigure[Reinforcement Learning]{
\begin{tabular}{l@{\hspace{8pt}}l}
\toprule
\bf Hyperparameter & \bf Value \\
\midrule
Learning rate & 5e-6 \\ 
Optimizer & AdamW \\ 
Temperature & 0.9 \\
Top-$P$ & 0.9 \\
Adv. estimator & GRPO \\
Clip ratio & 0.2 \\
\bottomrule
\end{tabular}
}
\label{tab:algorithms_specifications}
\end{table}

\subsection{Evaluation Metrics}


We use element-based and set-based validity, as well as element-based and
set-based generality, to evaluate the quality of grammars. In addition, we
employ \emph{effectiveness} to assess how well test cases generated by the
grammar distinguish incorrect solutions~\cite{sung2025logicase}.

The \emph{element-based effectiveness}~$E_\text{elt}$ measures the average ratio
of incorrect solutions distinguished by each test case to the total number of
incorrect solutions. In contrast, \emph{set-based effectiveness}~$E_\text{set}$
is the proportion of incorrect solutions that are distinguished by at least one
test case. Formally, given a set~$Y$ of incorrect solutions and a correct
solution~$\hat{y}$, each score for a set~$X$ of valid test cases is defined as
follows:
\begin{align*}
    E_{\text{elt}}(X) &= \mathbb{E}_{x \in X} \left[
        \mathbb{E}_{y \in Y} \left[ y(x) \ne \hat{y}(x) \right]
    \right], \\
    E_{\text{set}}(X) &= \mathbb{E}_{y \in Y} \left[
        \exists x \in X : y(x) \neq \hat{y}(x)
    \right].
\end{align*}

Both scores are zero if the set~$X$ contains any invalid test case. In our
evaluation, we generate 10 test cases from each grammar~$G$ to compute both
element-based and set-based effectiveness.

\subsection{Baseline Methods}

\subsubsection{Mutation-based fuzzing}

We adopt a mutation-based fuzzing strategy using both public and private test cases from the CodeContests dataset. Test cases are tokenized based on whitespace and newline delimiters. We randomly mutate 30\% of the tokens, with mutation strategies tailored to the token type---such as integers, floats, or strings. This targeted mutation preserves input semantics while enabling effective exploration of edge cases within the original specification constraints.

\subsubsection{Direct test case generation via LLMs}

We leverage four commercial LLMs for direct test case generation:
OpenAI's ChatGPT-4\footnote{\texttt{gpt-4-1106-preview}},
ChatGPT-4o\footnote{\texttt{gpt-4o-2024-08-06}},
Google's Gemini\footnote{\texttt{gemini-1.5-pro}} and
Anthropic's Claude-3.5\footnote{\texttt{claude-3-5-sonnet-20241022}}.
Each model is provided with a detailed problem specification and a strict output
format to ensure consistency. For each problem, we generate 10~test cases
directly based on the given specification. The following prompt is used for the
direct test case generation.



\begin{promptbox}[title=Prompt for Direct Test Case Generation]
\small
I will give you the specification of a problem.

\vspace*{0.3cm}

Generate 10 valid test cases for the below specification. 

Print all the testcase in the format below. 

Strictly, just print the test cases; do not
print anything else other than the test cases in a JSON format.

JSON format
should be: \{``testcases'': [``testcase1'', ``testcase2'', \ldots, ``testcase10'']\}

\vspace*{0.3cm}

Each test case should be in one line using \textbackslash n.

\{\{specification\}\}
\end{promptbox}





\subsubsection{Grammar-based test case generation via LLMs}


We leverage both the open-source and closed-source LLMs to evaluate the
performance of grammar-based test case generation. For closed-source models, we
evaluate ChatGPT, Gemini, Claude, DeepSeek, and, for open-source models, we
evaluate Gemma~\cite{DBLP:journals/corr/abs-2403-08295}, Qwen~\cite{DBLP:journals/corr/abs-2309-16609}, LLaMA~\cite{DBLP:journals/corr/abs-2407-21783}, StarChat~\cite{DBLP:journals/corr/abs-2402-19173},
Mistral~\cite{DBLP:journals/corr/abs-2310-06825} and DeepSeek~\cite{deepseek-r1}.

Each model is provided with the problem specification, accompanied by a detailed
5-shot CoT reasoning prompt, 18 instruction rules combined with the strict
output format ensuring the consistent generation of the corresponding~CCFG.
Table~\ref{tab:grammar_llms} summarizes the LLMs we used in our experiments for the
evaluation of grammar-based test case generation.

\begin{table}[htbp]
\caption{LLMs for Evaluation of Grammar-Based Test Case Generation}
\label{tab:grammar_llms}
\centering
\begin{tabular}{ll}
\toprule
\bf Source & \bf Model \\
\midrule
\multirow{4}{*}{Open} & ChatGPT-4, ChatGPT-4o \\
& Gemini-1.5-Pro \\
& Claude 3.5 Sonnet \\
& DeepSeek-V3, DeepSeek-R1 \\
\cmidrule{2-2}
\multirow{6}{*}{Closed} & Gemma2-9B \\
& Qwen 2.0-7B, Qwen 2.5-7B, Qwen 2.5-32B \\
& LLaMA 3.1-8B \\
& StarChat2-15B \\
& Mistral-Small-24B \\
& DeepSeek-V2-Lite-Base, DeepSeek-R1-Distill-Qwen-32B \\
\bottomrule
\end{tabular}
\end{table}


\section{Experimental Results and Analysis}



\begin{table}[t]
\centering
\setlength{\tabcolsep}{4pt}
\caption{
    Validity and effectiveness of test cases in dataset and two baseline
    methods: fuzzing and direct generation.
}
\label{tab:results}
\begin{tabular}{llcccc}
\toprule
\multirow{2}{*}{\vspace{-5pt}\bf Category} &
\multirow{2}{*}{\vspace{-5pt}\bf Type} &
\multicolumn{2}{c}{\bf Validity~(\%)} &
\multicolumn{2}{c}{\bf Effectiveness~(\%)} \\

\cmidrule(r){3-4}
\cmidrule{5-6}

& &
Elt.-based &
Set-based &
Elt.-based &
Set-based \\

\midrule

\multirow{3}{*}{Dataset} &
Public &
99.62 & 99.62 & 32.42 & 40.41 \\
& Private &
76.66 & 76.66 & 39.87 &  71.19\\
& Generated &
77.87 & 30.00 & 17.38 & 27.59 \\

\cmidrule{2-6}

\multirow{2}{*}{Fuzzing} &
Public &
80.96 & 47.77 & 19.09 & 30.43 \\
& Private &
63.77 & 31.48 & 16.78 & 27.54 \\

\cmidrule{2-6}

\multirow{4}{*}{Direct} &
ChatGPT-4 &
91.38 & 79.25 &  39.68 & 63.99 \\
& ChatGPT-4o &
83.03 & 71.11 &  35.39 & 58.01 \\
& Gemini &
80.07 & 61.11 & 28.37 & 44.96 \\
& Claude-3.5 &
89.03 & 84.44 & 38.35 & 65.05 \\
\bottomrule
\end{tabular}
\end{table}


\subsection{Overall analysis}

The experimental results clearly demonstrate the effectiveness of the proposed
\sysname~framework across multiple dimensions of test case quality. By integrating
supervised fine-tuning, RL with GRPO, \sysname~significantly outperforms both direct test case generation
methods and other grammar-based baselines. In particular, \sysname~achieves 96.66\%
set-based validity, 95.92\% set-based generality, and 80.67\% set-based
effectiveness, representing state-of-the-art performance on complex input
specifications for competitive programming problems. For reference, the
set-based effectiveness of ground-truth grammars is 83.70\%, which falls short
of 100\% due to the use of only 10 test cases per problem. This score improves
as more test cases are added.

Table~\ref{tab:main_results} shows that LLMs with few-shot prompts successfully
learn how to translate specifications into grammars, even with five
examples~(open- and closed-source). We can also observe that the performance of
an open-source LLM, DeepSeek-R1-Distill-Qwen, with 14B parameters dramatically
improves by supervised fine-tuning with less than 1,000 specification-grammar
pairs~(\sysname).

\begin{tcolorbox}[boxsep=1pt,left=4pt,right=4pt,top=4pt,bottom=4pt,boxrule=0pt]
{\bf Answer to RQ1.} LLMs can generalize effectively even with five example
prompts. Moreover, supervised fine-tuning with a small number of ground-truth
grammars can further improve performance.
\end{tcolorbox}

\begin{table*}[t]
\centering
\caption{Comparison of CCFG-based test case generation methods. For each metric, the best score is highlighted in bold, while the second-best is indicated with an underline.}
\label{tab:main_results}
\begin{tabular}{lllcccccc}
\toprule
\multirow{2}{*}{\bf \vspace{-5pt} Source} &
\multirow{2}{*}{\bf \vspace{-5pt} Prompting} &
\multirow{2}{*}{\bf \vspace{-5pt} Model} &
\multicolumn{2}{c}{\bf Validity~(\%)} &
\multicolumn{2}{c}{\bf Effectiveness~(\%)} &
\multicolumn{2}{c}{\bf Generality~(\%)} \\

\cmidrule(r){4-5}
\cmidrule(r){6-7}
\cmidrule(r){8-9}

& & & 
Elt.-based &
Set-based &
Elt.-based &
Set-based &
Elt.-based &
Set-based \\

\midrule


\multirow{14}{*}{Open-sourced} 
& \multirow{11}{*}{Single-Turn}& StarChat2-15B &
6.85 & 6.29 & 3.68 & 5.64 & 6.54 & 6.29 \\
&&LLaMA 3.1-8B &
13.66 & 12.96 & 7.05 & 11.37 & 11.64 & 44.48 \\
&& Mistral-Small-24B &
15.53 & 14.44 & 8.31 & 12.88 & 12.52 & 11.85 \\
&& Qwen 2.0-7B &
21.02 & 20.74 & 11.53 & 16.77 & 16.66 & 16.29 \\
&& Qwen 2.5-coder-7B &
39.47 & 37.77 & 18.52 & 31.53 & 36.26 & 35.55 \\
&& Qwen 2.5-32B &
75.67 & 71.85 & 30.16 & 60.91 & 72.34 & 71.11 \\
&& Gemma2-9B &
43.88 & 41.48 & 20.83 & 35.58 & 42.49 & 41.11 \\
&&DeepSeek-LLama-7B &
11.33 & 10.74 & 5.08 & 8.44 & 8.87 & 8.51 \\
&& DeepSeek-V2-Lite-Base &
32.13 & 30.37 & 15.84 & 26.38 & 31.43 & 30.37 \\
&& DeepSeek-Qwen-14B
&58.44 &57.03 &27.68 &47.31 &55.04 &53.7 \\
&& DeepSeek-Qwen-32B &
74.36 & 71.11 & 35.07 & 58.92 & 68.82 & 67.03 \\
\cmidrule{2-9}
&\multirow{3}{*}{Multi-Turn} 
&Qwen 2.5-32B & 78.54 & 75.55 & 37.89 & 63.20 & 74.65 & 73.33\\
&&DeepSeek-Qwen-14B
&69.58 &66.66 &32.37 &54.61 &63.47 &61.48 \\
&&DeepSeek-Qwen-32B
&80.10 &77.03 &38.71 &64.95 &74.15 &72.96 \\
\midrule

\multirow{11}{*}{Closed-sourced}
& \multirow{6}{*}{Single-Turn} 
& Gemini 1.5-Pro&
46.90 & 45.92 & 24.62 & 40.70 & 43.67 & 42.59 \\
&& ChatGPT-4 &
78.69 & 78.14 & 38.43 & 66.01 & 73.46 & 71.85 \\
&& ChatGPT-4o &
81.60  &80.00     &39.64  &66.42  &80.26  &80.00 \\
&& Claude 3.5 &
87.99 &87.40 &46.05 &74.64 &85.54 & 85.18 \\
&& DeepSeek-V3 &
82.65 & 81.11 & 40.81 & 68.08 & 77.67 & 77.40 \\
&& DeepSeek-R1 &
83.07 & 81.48 & 42.00 &69.80 & 77.95 & 77.4 \\
\cmidrule{2-9}
& \multirow{5}{*}{Multi-Turn} 
& ChatGPT-4o
&84.94 &83.33 &41.20 &69.63 &82.55 &82.22 \\
&& ChatGPT-4
& 86.08  &85.18 &42.85 &72.22 & 79.61 &77.70 \\
&& Claude 3.5
&\underline{92.27}  &\underline{91.85} & \underline{47.94} &\underline{77.39} &\underline{89.03} & \underline{88.14} \\
&& DeepSeek-V3
&86.05 & 84.07 &42.91 &72.87 &80.74 &80.37 \\
&& DeepSeek-R1
&88.66 &88.49 &44.59 &73.30 &82.52 &81.58 \\
\midrule

CCFGT5~\cite{sung2025logicase} &  Single-Turn & CodeT5 &82.15 & 80.74 & 42.10 &  68.33& 82.97 & 82.11\\
\midrule
\rowcolor{lightgray!50} SAGE (ours)& 
Single-Turn
& 
DeepSeek-R1-Distill-Qwen-14B
&
{\bf 97.28} & {\bf 96.66} & {\bf 50.66} & {\bf 80.67} & {\bf 96.75} & {\bf 95.92} \\
\midrule
Ground-truth & - & - & 100.00  & 100.00 & 52.61 & 83.70  & 100.00 & 100.00  \\
\bottomrule
\end{tabular}
\end{table*}


\subsection{Direct Generation vs. Grammar-Based Generation}

Direct test case generation methods, while simple and fast, often struggle to conform to the intricate constraints specified in problem descriptions. Grammar-based approaches address this limitation by explicitly modeling structural rules, enabling syntactically and semantically coherent test generation. The results in Tables~\ref{tab:results} and \ref{tab:main_results} indicate that grammar-based generation consistently outperforms direct generation in terms of both validity and effectiveness, especially when combined with fine-tuning and RL.

\begin{figure}[t]
  \centering
  \includegraphics[width=0.5\textwidth]{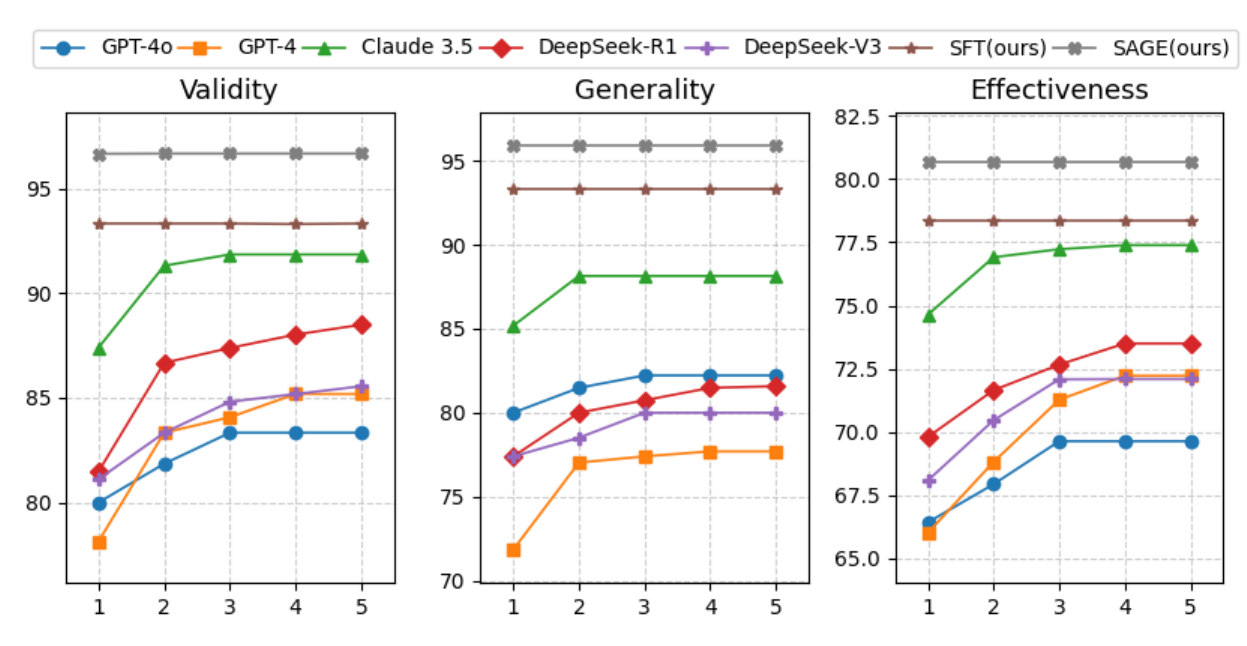}
  \caption{Changes in (set-based) validity, generality, and effectiveness over iterative refinement steps.}
  \label{fig:iteration}
\end{figure}


\subsection{Effectiveness of Prompting Strategy}

In our LLM-based grammar generation setup, we used a 5-shot example along with a chain-of-thought (CoT) style prompt and provided detailed rules for generating CCFG grammars.

\begin{promptbox}[title=Prompt for CCFG Generation from NL Specification]
\small

\begin{tcolorbox}[
  colframe=red,           
  boxrule=0.8pt,          
  sharp corners,          
  left=1pt, right=1pt,    
  top=1pt, bottom=1pt,    
  boxsep=0pt,             
  enhanced                
]
You are an expert grammar generator. 

\vspace*{0.1cm}

Strictly follow the given rules to convert specifications into grammars and constraints.

We will first give you the general rules and the examples of how the grammars and the constraints ....
\end{tcolorbox}
\hfill \textcolor{red}{Chain-of-Thought}

\vspace*{0.1cm}

\begin{tcolorbox}[
  colframe=red,           
  boxrule=0.8pt,          
  sharp corners,          
  left=1pt, right=1pt,    
  top=1pt, bottom=1pt,    
  boxsep=0pt,             
  enhanced                
]
General rules for generating the grammar:

1) \textsymb{<}S\textsymb{>} \textsymb{->} Start non-terminal: Represents the starting point of the grammar.

2) [variable] \textsymb{->} counter variable: Enclose variable names in brackets
to indicate they are counters variables and includes corresponding counter
non-terminals like \textsymb{<}T\_variable\textsymb{>} ....
\end{tcolorbox}
\hfill \textcolor{red}{Rule Instruction}

\vspace*{0.1cm}

\begin{tcolorbox}[
  colframe=red,           
  boxrule=0.8pt,          
  sharp corners,          
  left=1pt, right=1pt,    
  top=1pt, bottom=1pt,    
  boxsep=0pt,             
  enhanced                
]
The following are the examples for the generation of the grammar and constraint along with their reasons of how they are generated:

\textsymb{<}Specification\textsymb{>} Constraints \textbackslash n \textsymb{-}1000 \textsymb{<=} a, b \ldots \textsymb{<}/Specification\textsymb{>}
\textsymb{<}Reason\textsymb{>} ``The grammar begins with the start symbol \textsymb{<}S\textsymb{>}. Here, `N' is used as a counter variable for the array elements, ...." \textsymb{<}/Reason\textsymb{>}\\
\textsymb{<}Grammar\textsymb{>} ``\textsymb{<}S\textsymb{>} \textsymb{->} a \textsymb{<}s\textsymb{>} b" \textsymb{<}/Grammar\textsymb{>}

\textsymb{<}Constraint\textsymb{>} \\
``\textsymb{-}1000 \textsymb{<=} a \textsymb{<=} 1000", "\textsymb{-}1000 \textsymb{<=} b \textsymb{<=} 1000" \\
\textsymb{<}/Constraint\textsymb{>}

\ldots
\end{tcolorbox}
\hfill \textcolor{red}{Few-Shot Examples}
\vspace*{0.1cm}

\{\{specification\}\}

\vspace*{0.1cm}

Output in a JSON format. JSON format should be:

\{``grammar": \{ ``productions": [``"], ``constraints": [""]\}\}.
\end{promptbox}



Experimental results shown in Table~\ref{tab:prompting} justify our prompt design choices. As shown in the results, the proposed prompt configuration achieved the best performance for both ChatGPT-4 and Claude 3.5. Among the three prompt components---few-shot examples, rules, and CoT---their relative impact on performance can be ranked as follows: (1) few-shot examples $>$ (2) rules $>$ (3) CoT. Interestingly, the results indicate that even when the grammar generation rules were removed, the model still achieved the second-highest performance among all variants. While increasing the number of examples might further improve performance, we limited it to five due to the prompt length constraint. Notably, in the case of SFT models, the rule component was excluded from the input prompt. This was because, through supervised learning and RL, the model was able to implicitly learn the grammar construction rules without requiring explicit rule descriptions.


\subsection{Effectiveness of RL}

Table~\ref{tab:ablation} shows that incorporating GRPO with reward functions based on grammar validity and generality leads to consistent improvements over the supervised-only baseline.

\begin{table}[t]
\centering
\setlength{\tabcolsep}{3pt}
\caption{Ablation study on the impact of each proposed component.}
\label{tab:ablation}
\begin{tabular}{lcccccc}
\toprule
\multirow{2}{*}{\vspace{-4pt}\bf Method} &
\multicolumn{2}{c}{\bf Validity} &
\multicolumn{2}{c}{\bf Effectiveness} &
\multicolumn{2}{c}{\bf Generality} \\
\cmidrule(r){2-3}
\cmidrule(r){4-5}
\cmidrule{6-7}
& Elt. & Set & Elt. & Set & Elt.  & Set \\
\midrule
 Base Model  
&58.44& 57.03 & 27.68 & 47.31 & 55.04 & 53.70 \\ \cdashline{1-7} \noalign{\vskip 3pt}
\quad (+) Supervised Fine-Tuning &
94.91 & 93.33 & 49.19 & 78.38 & 94.34 & 93.33 \\ 
 \qquad (+) GRPO Learning  &
{\bf 97.28} & {\bf 96.66} & {\bf 50.66} & {\bf 80.67} & {\bf 96.75} & {\bf 95.92} \\
\bottomrule
\end{tabular}
\end{table}

\begin{table}[t]
\centering
\setlength{\tabcolsep}{2.5pt}
\caption{Ablation study on the impact of each component in the prompt. For each model and metric, the best score is highlighted in bold, while the second-best is indicated with an underline.}
\label{tab:prompting}
\begin{tabularx}{\linewidth}{lcccYYYYYY}
\toprule
\multirow{2}{*}{\vspace{-4pt}\bf Method} & 
\multirow{2}{*}{\vspace{-4pt}\bf Shots} &
\multirow{2}{*}{\vspace{-4pt}\bf Rule} & 
\multirow{2}{*}{\vspace{-4pt}\bf CoT} & 
\multicolumn{2}{c}{\bf Validity} & 
\multicolumn{2}{c}{\bf Effectiveness} & 
\multicolumn{2}{c}{\bf Generality} \\
\cmidrule(r){5-6}
\cmidrule(r){7-8}
\cmidrule{9-10}
& & & &  Elt. & Set & Elt. & Set & Elt. & Set \\
\midrule
\multirow{8}{*}{ChatGPT-4} & \multirow{4}{*}{1}  & \greencmark & \greencmark &63.71 & 62.96 & 31.75 & 53.25 & 59.02 & 58.14 \\
&  &\redxmark & \greencmark & 37.08 & 37.03 & 20.46 & 32.81 & 34.42 & 34.07 \\
&   & \greencmark & \redxmark & 55.81 & 55.55 & 26.85 & 46.72 & 53.01 & 52.59 \\
&  &\redxmark & \redxmark & 23.48 & 23.33 & 12.52 & 26.38 & 21.36 & 21.11 \\\cmidrule{2-10}
& \multirow{4}{*}{5} & \greencmark & \greencmark  & {\bf 78.69} & {\bf 78.14} & {\bf 38.43} & {\bf 66.01} & {\bf 73.46} & {\bf 71.85} \\
&  &\redxmark & \greencmark & \underline{72.62} & \underline{72.22} & \underline{35.55} & \underline{61.09} & \underline{68.35} & \underline{67.40} \\
&  & \greencmark & \redxmark & 72.32 &  71.22 & 37.90 & 65.39 & 70.51 & 69.44 \\
&  &\redxmark & \redxmark & 64.85 & 64.07 & 32.03 & 54.56 & 56.85 & 55.92 \\
\midrule
\multirow{8}{*}{Claude 3.5} & \multirow{4}{*}{1} & \greencmark & \greencmark & 85.06 & 82.96 & 43.12 & 70.09 & \underline{85.18} & \underline{84.44} \\
& & \redxmark & \greencmark & 38.36 & 37.77 & 18.87 & 31.25 & 34.18 & 33.70 \\
& & \greencmark & \redxmark & 51.23 & 47.77 & 23.58 & 38.23 & 51.79 & 50.74 \\
& & \redxmark  & \redxmark & 36.60 & 35.92 & 17.36 & 30.33 & 34.43 & 34.07 \\\cmidrule{2-10}
& \multirow{4}{*}{5} & \greencmark & \greencmark  & {\bf 87.99} & {\bf 87.40} & {\bf 46.05} & {\bf 74.64} & {\bf 85.54} & {\bf 85.18} \\
& & \redxmark & \greencmark & \underline{85.49} & \underline{85.18} & \underline{44.03} & \underline{72.65} & 82.83 & 82.22\\
& & \greencmark & \redxmark & 83.47  & 82.96 & 42.47 & 70.76  & 80.74 & 80.37 \\
& & \redxmark & \redxmark  & 83.27 & 82.59 & 42.28 & 70.73 & 81.73 & 81.11 \\

\bottomrule
\end{tabularx}
\end{table}





Removing GRPO training leads to a more noticeable drop in all metrics, especially in generality, demonstrating that RL is essential for aligning grammars with structural and semantic constraints.
Removing supervised fine-tuning yields the most drastic degradation across all metrics, suggesting that SFT is foundational for initial alignment of the LLM to the specification-to-grammar task.

Our ablation study validates the effectiveness of our stepwise architecture and shows that each component---SFT, RL, and iterative feedback---plays a complementary and essential role in producing high-quality CCFGs.

\begin{tcolorbox}[boxsep=1pt,left=4pt,right=4pt,top=4pt,bottom=4pt,boxrule=0pt]
{\bf Answer to RQ2.} RL enables the model to refine grammar candidates by exploiting structured reward signals, achieving higher set-based validity and generality across unseen specifications.
\end{tcolorbox}

\subsection{Single-Turn vs. Multi-Turn Prompting}



Figure~\ref{fig:iteration} shows the trends of validity, generality, and effectiveness over iterative refinements. We can see that the iterative feedback shows clear benefits for models that are not fine-tuned, leading to consistent improvements in grammar generation quality across multiple iterations. However, iterative feedback does not improve our fine-tuned models. This is because out of 271 benchmark specifications, our model produced ill-formed grammars in only three cases, indicating that the model already generates well-formed grammars with high probability.



\begin{tcolorbox}[boxsep=1pt,left=4pt,right=4pt,top=4pt,bottom=4pt,boxrule=0pt]
{\bf Answer to RQ3.} Iterative refinement helps closed-source LLMs recover from early grammar generation errors and produce more syntactically and semantically coherent grammars. In contrast, \sysname~already produces well-formed grammars in most cases, which limits the marginal benefit of iterative feedback process.
\end{tcolorbox}

 
\subsection{Case Study}

We analyzed scenarios in which our iterative approach struggles to generate
accurate CCFGs and corresponding constraints. A recurring failure pattern
emerges when the model fails to address issues related to the well-formedness of
the grammar. This deficiency causes stagnation in the validity score and hinders
further progress.

\begin{tcolorbox}[boxsep=1pt,left=4pt,right=4pt,top=4pt,bottom=4pt,boxrule=0pt,breakable]
\begin{example}[Input Specification: 1419\_A. Digit Game]\label{example:specification_case_study_validity_1}
\hfill
\begin{itemize}[leftmargin=1.0em,itemsep=-0.2em]
    \item The first line of each match description contains an integer~$n$~($1
    \le n \le 10^3$)---the number of digits of the generated number.
    \item The second line of each match description contains an n-digit positive integer without leading zeros.
\end{itemize}
\end{example}

\textbf{Model Output:}
\begin{align*}
    S &\to [t]\ \textsymb{<n>}\ T_t, \\[-2pt]
    T_i &\to T_{i-1}\ \textsymb{<n>}\ [n]\ \textsymb{<n>}\ \textsymb{[1-9]\{1\}} \\[-2pt]
        &\quad\;\; \textsymb{<s>}\ [0-9][0-9]\{n-1\}, \\[-2pt]
    T_1 &\to \ [n]\ \textsymb{<n>}\ [1-9]\{1\}\ \textsymb{<s>}\ [0-9]\textsymb{\{\ensuremath{n{-}1}\}}
\end{align*}

\vspace{0.5em}

\textbf{Validator Feedback:} Counter operator parse failed: $n-1$ and too many nodes found, not valid grammar.
\end{tcolorbox}

In Example~\ref{example:specification_case_study_validity_1}, the iterative process failed to resolve a node overflow error and incorrectly handled the symbolic
counter expression \textsymb{n-1}. Specifically, the model produced an invalid grammar
using expressions like \textsymb{[0-9]\{n-1\}}, which are syntactically unsupported due
to the improper use of dynamic counters. The correct formulation should have
employed a valid counter structure, such as \textsymb{<S>} $\to$ \textsymb{[t]}
\textsymb{<n>} \textsymb{<T\_t>}, \textsymb{<T\_i>} $\to$
\textsymb{<T\_i-1>} \textsymb{<n>} \textsymb{[n]} \textsymb{<n>} \textsymb{[0-9]\{n\}}, and
\textsymb{<T\_1>} $\to$ \textsymb{[n]} \textsymb{<n>} \textsymb{[0-9]\{n\}}, where the repetition
count remains both syntactically valid and grounded in the defined non-terminal
values.

\begin{tcolorbox}[boxsep=1pt,left=4pt,right=4pt,top=4pt,bottom=4pt,boxrule=0pt,breakable]
\begin{example}[Input Specification: 453\_E. Little Pony and Lord Tirek Game]
\label{example:specification_case_study_generality_1}
\hfill
\begin{itemize}[leftmargin=1.0em,itemsep=-0.2em]
    \item The first line contains an integer n ($1 \le n \le 10^5$)---the number
    of ponies. Each of the next $n$~lines contains three integers $s_i$, $m_i$,
    $r_i$~($0 \le s_i \le m_i \le 10^5; 0 \le r_i \le 10^5$), describing a pony.
    \item The next line contains an integer $m~(1 \le m \le 10^5)$---the number
    of instructions.
    \item Each of the next $m$~lines contains three integers $t_i$, $l_i$,
    $r_i$~($0 \le t_i \le 10^9; 1 \le l_i \le r_i \le n$), describing an
    instruction of Lord Tirek.
\end{itemize}
\end{example}

{\bf Model Output: }
\begin{align*}
    1 \le n \le 10^5, 0 \le s_i \le m_i \le 105, 0 \le r_i \le 105, \\[-2pt]
    1 \le m \le 10^5, 0 \le t_i \le 10^9, 1 \le l_i \le r_i \le n
\end{align*}
\end{tcolorbox}

In Example~\ref{example:specification_case_study_generality_1}, the grammar
failed to capture generality due to a misinterpretation of the variables
\textsymb{m\_i} and \textsymb{r\_i}, where $10^5$ was mistakenly interpreted as
$105$. This misinterpretation led to the generation of ill-formed grammars.

On the other hand, Figure~\ref{fig:iteration_overview} illustrates a successful progression of our iterative framework. Initially, the model generated a grammar with an invalid regex repetition pattern, such as \textsymb{[0-9]+}, which violates the syntactic constraints of the CFG formalism. In the subsequent turn, the model overcorrected by unrolling the repetition into \textsymb{[0-9][0-9][0-9]}, which led to node explosion and violated structural constraints. In the final iteration, the model successfully abstracted the numeric token as a symbolic variable \textsymb{a\_i}, producing a well-formed grammar: \textsymb{<S>} $\to$ \textsymb{[n]} \textsymb{<n>} \textsymb{<T\_n>}, \textsymb{<T\_i>} $\to$ \textsymb{<T\_i-1>} \textsymb{<s>} \textsymb{a\_i}, and \textsymb{<T\_1>} $\to$ \textsymb{a\_i}.

\section{Related Work}

\subsection{Test Case Generation with LLMs}


Recent advancements in LLMs such as Codex, ChatGPT, and Gemini have transformed automated test case generation.. Approaches like TestAug~\cite{YangHS0022}, CodeT~\cite{DBLP:conf/iclr/ChenZNZLLC23}, ChatTESTER~\cite{Yuan0DW00L24}, TestBench~\cite{DBLP:journals/corr/abs-2409-17561,DBLP:conf/icse/NieBLMG23}, TestGen-LLM~\cite{DBLP:conf/sigsoft/AlshahwanCFGHHM24}, TestPilot~\cite{DBLP:journals/corr/abs-2302-06527}, TestSpark~\cite{DBLP:conf/icse/SapozhnikovOPKD24}, TestEval~\cite{WangYWHCSZCM24}, and EffiBench-X~\cite{qing2025effibenchxmultilanguagebenchmarkmeasuring} employ LLMs to directly generate test inputs from problem descriptions, often leveraging few-shot prompting or CoT reasoning to guide generation. These approaches are attractive due to their simplicity and ease of deployment, producing syntactically valid examples without manual grammar engineering. 
 Furthermore, ChatUniTest~\cite{DBLP:conf/sigsoft/ChenHZHDY24} and ChatTester~\cite{DBLP:journals/pacmse/Yuan0DW00L24} generate test cases through an iterative refinement process to progressively improve the quality and correctness of the outputs.

However, existing methods face two critical limitations: (1) generated test cases frequently violate complex or nested input constraints, and (2) they often lack diversity and fail to capture edge cases. These limitations reduce their effectiveness in demanding settings such as competitive programming or program repair.

Our work addresses these issues by moving beyond direct example generation. Instead, we use LLMs to generate grammars that encode the underlying structure of valid inputs, and we iteratively refine these grammars using well-formedness, validity, and generality metrics. This structured pipeline enhances the reliability, interpretability, and robustness of test case generation compared to purely example-driven techniques.

\subsection{Grammar-Based Test Case Generation and Fuzzing}


Grammar-based test case generation has long been a cornerstone of software testing, particularly in fuzzing scenarios where structured input formats are required. Traditional fuzzing tools like AFL~\cite{afl} and early
grammar-based systems~\cite{godefroid2008grammar} rely on manually crafted CFGs to ensure syntactic correctness during test input generation. More recently, grammar-aware fuzzers such as Superion~\cite{DBLP:conf/icse/Wang0WL19} and Gramatron~\cite{SrivastavaP21} and neural-guided approaches like CodaMosa~\cite{LemieuxILS23}, have demonstrated that structural guidance enables more effective exploration of program behavior. In parallel, mutation-based and structure-preserving fuzzers---including FuzzIL~\cite{Gro2018FuzzILCG}, and and approaches based on code fragments  Ifuzzer~\cite{DBLP:conf/esorics/VeggalamRHB16} have shown that leveraging evolutionary strategies or intermediate representations can further enhance fuzzing effectiveness.
Additionally, tools like EvoSuite~\cite{DBLP:conf/sigsoft/FraserA11} start from random test suites and iteratively mutate them to maximize code coverage.

Nonetheless, a major limitation of grammar-based techniques is their reliance on human-written grammars, which are labor-intensive to construct and often fail to generalize across diverse input formats---particularly in competitive programming, where each problem presents a new specification. To address this, we leverage a fine-tuned LLM to automatically generate CCFGs from problem descriptions. Moreover, we augment static generation with a dynamic feedback loop that iteratively refines grammar quality, improving the well-formedness, validity, and generality over multiple iterations.

\subsection{RL with Verifiable Rewards (RLVR)}

RL has become a central paradigm for aligning LLMs with complex objectives~\cite{RLHF,DPO,deepseek-r1}. In particular, instruction-tuning and RL with human feedback~(RLHF) have shown impressive results in aligning models with human preferences~\cite{RLHF}. However, in domains such as code generation or grammar synthesis that require strict semantic or syntactic correctness, relying on human feedback may be insufficient. Instead, we need verifiable reward signals for guiding the learning process toward semantic and syntactic correctness.

Recent works in code and synthesis domains have started to embrace this idea of using verifiable reward functions. Mercury~\cite{DuLJLN24} introduces RL to improve code execution efficiency by fine-tuning CodeLLMs, using execution metrics as reward signals. AceCoder~\cite{acecoder25} formulates RL with test case success as a verifiable reward and further enhances learning via automated test case synthesis and preference modeling. EffiBench~\cite{NEURIPS2024_15807b6e} proposes benchmark tasks for evaluating RLHF in program synthesis, focusing on measurable properties like latency and correctness. Most recently, Absolute Zero~\cite{zhao2025absolutezeroreinforcedselfplay} introduces a self-play strategy for code generation where correctness is determined via execution-based verification, offering a clean form of RL with directly observable rewards.

Our work builds upon this direction of RLVR, where reward signals can be computed and strongly correlated with objective quality metrics. In the context of grammar learning for test case generation, we define verifiable rewards in terms of validity and generality---two properties that are directly measurable and interpretable.

\section{Conclusions and Future Work}

In this paper, we introduced \sysname, a grammar-based test case generation framework that combines supervised fine-tuning and RL with verifiable rewards to address the challenges of generating valid and general test inputs from natural language specifications. By leveraging CCFGs, our approach systematically encodes structural and semantic constraints, enabling robust and scalable test case generation for competitive programming problems.

Our experiments show that \sysname~ significantly outperforms prior methods---including direct generation, mutation-based fuzzing, and grammar-based generation using both open- and closed-source LLMs---in terms of validity, generality, and effectiveness. In particular, the integration of GRPO with verifiable reward signals based on grammar validity and generality proved essential for improving performance across diverse problem settings. Additionally, our iterative refinement strategy led to substantial gains over single-turn generation, demonstrating the value of multi-step self-correction in grammar induction. 

As future work, we plan to incorporate incorrect or buggy reference code to guide the generation of targeted test cases, extend \sysname~ to more expressive grammar representations, and explore effectiveness-guided RL even with unlabeled specifications. We also aim to apply \sysname~ to real-world tasks such as program synthesis and automated grading, where precise input modeling is critical.

\newpage

\end{document}